\documentclass[11pt,letterpaper]{article}
\usepackage[letterpaper]{geometry}
\usepackage{acl2012}
\usepackage{times}
\usepackage{latexsym}
\usepackage{amsmath}
\usepackage{multirow}
\usepackage{url}
\usepackage{color}
\usepackage{xcolor}
\usepackage{graphicx}
\usepackage{subcaption}
\usepackage{tikz-dependency}
\usepackage{filecontents}
\usepackage{verbatim}
\usepackage{amsfonts}
\usepackage{pgfplots}
\usepackage{pgfplotstable}
\usepackage{nicefrac}
\pgfplotsset{width=7cm,compat=1.8}
\usepackage{etoolbox}
\makeatletter
\patchcmd\@combinedblfloats{\box\@outputbox}{\unvbox\@outputbox}{}{%
   \errmessage{\noexpand\@combinedblfloats could not be patched}%
}%

\newcommand{\@BIBLABEL}{\@emptybiblabel}
\newcommand{\@emptybiblabel}[1]{}
\newcommand{\ignore}[1]{}

\makeatother
\usepackage[hidelinks]{hyperref}

\title{Scheduled Multi-Task Learning: From Syntax to Translation}

\author{Eliyahu Kiperwasser\thanks{~~Work carried out during summer internship at IBM Research.} \\
Computer Science Department \\
  Bar-Ilan University \\
  Ramat-Gan, Israel \\
  {\tt elikip@gmail.com} \\\And
  Miguel Ballesteros \\
  IBM Research \\
  1101 Kitchawan Road, Route 134 \\
  Yorktown Heights, NY 10598. U.S \\
  {\tt miguel.ballesteros@ibm.com} \\}

\date{}

\begin{document}
\maketitle
\begin{abstract}
  Neural encoder-decoder  models  of  machine translation  have  achieved  impressive  results, while learning linguistic knowledge of both the source and target languages in an implicit end-to-end manner. We propose a framework in which our model begins learning syntax and translation interleaved, gradually putting more focus on translation. Using this approach, we achieve considerable improvements in terms of BLEU score on relatively large parallel corpus (WMT14 English to German) and a low-resource (WIT German to English) setup.
\end{abstract}

\section{Introduction}

Neural Machine Translation (NMT) \cite{DBLP:conf/emnlp/KalchbrennerB13,DBLP:conf/nips/SutskeverVL14,DBLP:journals/corr/BahdanauCB14} has recently become the state-of-the-art approach to machine translation \cite{DBLP:conf/wmt/BojarCFGHHJKLMN16}. One of the main advantages of neural approaches is the impressive ability of RNNs to act as feature extractors over the entire input \cite{DBLP:journals/tacl/KiperwasserG16}, rather than focusing on local information. Neural architectures are able to extract linguistic properties from the input sentence in the form of morphology \cite{belinkov-EtAl:2017:Long} or syntax \cite{TACL972}.

Nonetheless, as shown in \newcite{dyer-EtAl:2016:N16-1} and \newcite{dyer:2017:CoNLL}, systems that ignore explicit linguistic structures are incorrectly biased and they tend to make overly strong linguistic generalizations. Providing explicit linguistic information \cite{dyer-EtAl:2016:N16-1,kuncoro-EtAl:2017:EACLlong,DBLP:conf/wmt/NiehuesC17,DBLP:conf/wmt/SennrichH16,eriguchi-tsuruoka-cho:2017:Short,DBLP:conf/acl/AharoniG17a,DBLP:journals/corr/NadejdeRSDJKB17,DBLP:conf/emnlp/BastingsTAMS17,matthews18naacl} has proven to be beneficial, achieving higher results in language modeling and machine translation.

Multi-task learning (MTL) consists of being able to solve synergistic tasks with a single model by jointly training multiple tasks that look alike. The final dense representations of the neural architectures encode the different objectives, and they leverage the information from each task to help the others. For example, tasks like multiword expression detection and part-of-speech tagging have been found very useful for others like combinatory categorical grammar (CCG) parsing, chunking and super-sense tagging \cite{bingel-sogaard:2017:EACLshort}.

In order to perform accurate translations, we proceed by analogy to humans. It is desirable to acquire a deep understanding of the languages; and, once this is acquired it is possible to learn how to translate gradually and with experience (including revisiting and re-learning some aspects of the languages). We propose a similar strategy by introducing the concept of Scheduled Multi-Task Learning (Section \ref{sec:smtl}) in which we propose to interleave the different tasks. 

In this paper, we propose to learn the structure of language (through syntactic parsing and part-of-speech tagging) with a multi-task learning strategy with the intentions of improving the performance of tasks like machine translation that use that structure and make generalizations. We achieve considerable improvements in terms of BLEU score on a relatively large parallel corpus (WMT14 English to German) and a low-resource (WIT German to English) setup. Our different scheduling strategies show interesting differences in performance both in the low-resource and standard setups.

\section{Sequence to Sequence with Attention}
\label{seq2seq}
Neural Machine Translation (\textit{NMT})  \cite{DBLP:conf/nips/SutskeverVL14,DBLP:journals/corr/BahdanauCB14} directly models the conditional probability $p(y|x)$ of the target sequence of words  \mbox{$ y = <y_1, \dots , y_T> $} given a source sequence \mbox{$ x = <x_1, \dots , x_S> $}. In this paper, we base our neural architecture on the same sequence to sequence with attention model; in the following we explain the details and describe the nuances of our architecture.

\subsection{Encoder}
We use bidirectional LSTMs to encode the source sentences \cite{DBLP:series/sci/2012-385}. Given a source sentence \mbox{$ x = <x_1, \dots , x_m> $}, we embed the words into vectors through an embedding matrix $W^S$, the vector of the $i$-th word is $W^S  x_i$. We get the  representations  of  the $i$-th word by  summarizing  the  information of  neighboring words using bidirectional LSTMs \cite{DBLP:journals/corr/BahdanauCB14},
\begin{align}
h^F_i = LSTM^F(h^F_{i-1},\ W^S  x_i) \\
h^B_i = LSTM^B(h^B_{i+1},\ W^S  x_i).
\end{align}

The forward and backward representation are concatenated to get the bi-directional encoder representation  of  word $i$ as $h_i = [h^F_{i}, h^B_{i}]$.

\subsection{Decoder}
The decoder generates one target word per time-step, hence, we can decompose the conditional probability as
\begin{align}
\log\ p(y|x) = \sum_j p(y_j | y_{<j}, x).
\end{align}


The decoding procedure consists of two main processes: attention and LSTM based decoding. The attention mechanism calculates the weights ($\alpha_i$) for each source word based on the words translated/decoded so far. The model gives higher weight to words that are more relevant to decode the next word in the sequence. This is based on the words decoded so far represented by the decoder state ($d_j$), and the encoder representation of the sentence ($h_i$). Concretely, we use dot attention \cite{DBLP:conf/emnlp/LuongPM15} to calculate the attention weights. More formally, $\alpha_i$ is calculated as follows:
\begin{align}
& e_i = d_j^\top h_{i} \\ 
& \alpha_i = \frac{\exp(e_i)}{\sum_k \exp(e_k)}.
\end{align}

A vector representation ($c_j$) capturing the information relevant to this time-step is computed by a weighted sum of the encoded source vector representations using $\alpha$ values as weights.
\begin{align}
c_j = \sum_i \alpha_i \cdot h_i.
\end{align}

Given the sentence representation produced by the attention mechanism ($c_j$) and the decoder state capturing the translated words so far ($d_j$), the model decodes the next word in the output sequence. The decoding is done using a multi-layer perceptron which receives $c_j$ and $d_j$ and outputs a score for each word in the target vocabulary:
\begin{align}
& g_j = \tanh( W^1_{Dec}  d_j + W^1_{Att}  c_j  ) \\
& u_j = \tanh( g_j + W^2_{Dec}  d_j + W^2_{Att}  c_j  ) \\
& p(y_j | y_{<j}, x) \approx softmax( W_{out}  u_j + b_{out} ) .
\end{align}

\section{Many Tasks One Sequence to Sequence}
\label{sec:MTOSS}
Sequence to sequence models have been used for many tasks such as: machine translation \cite{DBLP:conf/nips/SutskeverVL14,DBLP:journals/corr/BahdanauCB14}, summarization \cite{DBLP:conf/emnlp/RushCW15} and syntax \cite{DBLP:conf/nips/VinyalsKKPSH15}. Several recent works have shown that parameter sharing between multiple sequence to sequence models that aim to solve different tasks may improve the accuracy of the individual tasks \cite[inter-alia]{DBLP:journals/corr/KaiserGSVPJU17,DBLP:journals/corr/LuongLSVK15,DBLP:conf/naacl/ZophK16,DBLP:conf/wmt/NiehuesC17,bingel-sogaard:2017:EACLshort}. 

We apply a simple yet effective approach to learn multiple tasks using a single sequence to sequence model inspired by \newcite{DBLP:journals/tacl/AmmarMBDS16}. All tasks share a common output vocabulary and generate terms according to (3). We learn multiple tasks simultaneously by prepending a special task embedding vector to the target. The task vector symbolizes the task we are focusing on. The model can solve each of the tasks it was trained on by priming the decoder with the token of each task. \newcite{DBLP:journals/tacl/JohnsonSLKWCTVW17} suggested to prepend a special embedding vector according to the desired target language. In contrast to our approach, they prepend the vector to the encoder and not to the decoder. 

We apply this methodology to jointly learn the multiple tasks, however many of the tasks are not of sequential nature (such as dependency parsing in which the output should be a well-formed dependency tree \cite{hudson1984word,melʹvcuk1988dependency}). We fit those into our sequence to sequence model in order to enrich the representation of other tasks, and increase the potential  information flow between the tasks. In what follows, we show which tasks (and how we linearize them) we solve jointly using our model and how we apply sequence to sequence modeling to those tasks.

\paragraph{Part-Of-Speech Tagging} Given a sentence and its part-of-speech annotation, we convert the task to translating between the sentence (as the source sequence) and the given part-of-speech tags as the target. A similar approach was suggested by \newcite{DBLP:conf/wmt/NiehuesC17}.

\begin{figure}
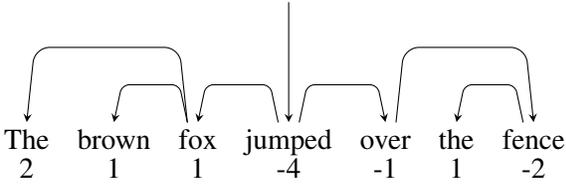

\begin{dependency}[hide label]
\begin{deptext}[column sep=0.2cm]
The \& brown \& fox \& jumped \& over \& the \& fence \\
2 \& 1 \& 1 \& -4 \& -1 \& 1 \& -2 \\
\end{deptext}
\depedge{3}{1}{det}
\depedge{4}{3}{nsubj}
\depedge{3}{2}{amod}
\deproot{4}{root}
\depedge{4}{5}{prep}
\depedge{5}{7}{pobj}
\depedge{7}{6}{det}
\end{dependency}
\caption{Illustration of the encoding of an unlabeled parsing tree into a sequence of distances. The first row contains the sentence (source) and its parse tree, and the second row contains the matching distances sequence (target). }
\label{fig:distances}
\end{figure}

\paragraph{Unlabeled Dependency Parsing} An unlabeled dependency tree annotation can be viewed as a sequence of heads, where for every node there is a unique incoming edge, that is, a single matching head. We convert the tree by scanning the sentence from left to right, and outputting the distance of each word to its head. We then convert the task to translating between the original sentence and the resulting sequence describing the unlabeled dependency tree (See Figure \ref{fig:distances}). Sequence of distances is an invertible representation of the sequence of heads, which is equivalent to an unlabeled tree. In contrast to a sequence of heads, learning a sequence of distances is able to generalize to sentences of arbitrary length (including length which are not seen or rarely seen in the training corpus). Distance to the syntactic heads has also been shown to be an effective feature when parsing sentences \cite{DBLP:conf/acl/McDonaldCP05}.

\paragraph{Predicting Dependency Relations-Labeled Dependency Parsing} Similarly to the conversion of the unlabeled dependency tree to a sequence, we scan all the words in the sentence from the beginning to the end. For each word encountered, we output the label of the dependency arc connecting it with its matching head word. We, therefore, learn to translate between the original sentence and the resulting sequence of dependency labels.

\paragraph{Machine Translation} Similarly to \newcite{DBLP:conf/nips/SutskeverVL14} and \newcite{DBLP:journals/corr/BahdanauCB14}, we use sequence to sequence to translate between a sentence written in a source language and a sentence written in a target language.

\section{Scheduled Multi-Task Learning}
\label{sec:smtl}

\begin{figure}
\centering
\includegraphics[width=0.5\textwidth]{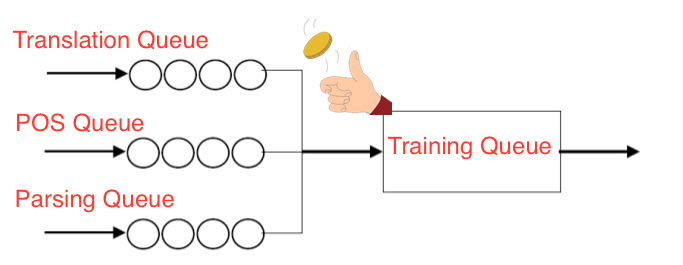}
\caption{Illustration of Scheduled Multi-Task Learning approach. This figure contains a flow chart describing how the training examples from multiple sources are gathered into a single training queue. The chart starts from multiple queues, each containing training examples belonging to different tasks (possibly from different datasets). Using a coin toss we choose the next queue to take the following training example from. The probability of each of the queues to be selected is determined by the scheduler.} 
\label{fig:SMTL}
\end{figure}

In order to produce accurate translations, neural machine translation systems have to learn syntax in order to generate grammatically correct sentences. Furthermore, translation systems have to disambiguate different parts-of-speech on the source side sentence, since a different part-of-speech can result in different translations. 
There are many sets of parameters able to capture the training data when employing LSTM (RNN) models. This applies to sequence to sequence models with attention. Each set of parameters provides a different level of generalization \cite{DBLP:conf/emnlp/ReimersG17}.
As suggested by \newcite{dyer:2017:CoNLL}, representations learned by the network do not capture the linguistic properties, and they are biased to make overly strong linguistic generalizations. 

Providing ``guidance'' to the sequence to sequence network at the beginning focusing it on a representation enriched with linguistic knowledge, such as syntax or part-of-speech tagging, helps it obtain information necessary for converging to a more general solution. We suggest interleaving the learning of the syntax and translation tasks, and gradually decrease the weight of the syntactically oriented tasks (auxiliary tasks). This enables the model to forget about the syntax examples and to put more focus on fitting the translation task as the training progresses.

Our approach, Scheduled Multi-Task Learning (SMTL), is a semi-supervised learning approach that generalizes the above scheme. Scheduled Multi-Task Learning continuously interleaves between three well-known previous methods: Multi-task learning, Pre-training, and Fine-tuning. 

Multi-Task Learning (\textit{MTL}) \cite{DBLP:journals/ml/Caruana97} solves synergistic tasks while maximizing the number of shared parameters. Sharing parameters for multiple tasks may increase the accuracy in tests for the individual tasks, thanks to representation bias which captures a more regularized representation fitted to multiple tasks \cite{DBLP:journals/jair/Baxter00} and using information from one task as hints to the other tasks \cite{DBLP:journals/jc/Abu-Mostafa90}. In case of independence between the features of the multiple tasks learned, we assume that enforcing the representation to accommodate multiple tasks can result in a drop in accuracy compared to the accuracy of each task learned separately \cite{DBLP:journals/ml/Caruana97,bingel-sogaard:2017:EACLshort}.

Pre-Training \cite{DBLP:journals/jmlr/CollobertWBKKK11} is a widely used approach \cite{goldberg2017neural} which initializes the parameters with the parameters used to solve a somewhat related task. Similarly, Fine-Tuning uses a small annotated in-domain corpus and a large annotated out-of-domain corpus to estimate parameters. We first learn using the large out-of-domain corpus and once that is finished, we continue learning (fine-tuning) on the in-domain corpus.
This is a common approach for transfer learning \cite{DBLP:conf/nips/YosinskiCBL14}. A related approach is to start with a pre-trained neural network model and fine-tune only the final layers in order to keep the coarse features detected for the previous task \cite{hinton2006reducing,DBLP:journals/jmlr/ErhanBCMVB10}.
Both approaches, facilitate encoding useful information from related tasks (Pre-training) or data-sets (Fine-tuning) without demanding that the representation accommodate both tasks, and can be viewed as regularization \cite{DBLP:journals/ml/Caruana97}.

Our Scheduled Multi-Task Learning approach unifies the above methods into a single framework. This framework contains multiple queues, where each queue contains the training examples belonging to a specific pair of tasks and datasets. In order to pick the next training example, we stochastically pick a queue ($q$) with time-dependent probability ($p^t_{q}$) and then we get the next example from the chosen queue (Figure \ref{fig:SMTL}). 

The probabilities ($p^t_q$) change as the training progresses according to a \textit{Schedule}. The Schedule could, for example, give a high probability at the beginning of the training process to some task (e.g. part-of-speech tagging) and gradually decrease the probability in favor of another task (e.g. translation). The latter schedule resembles the pre-training approach at the beginning by, later in the process, progressing to multi-task learning approach. Such a schedule enables harnessing hints from related tasks and also enforces a soft representation bias at the beginning of the training. This contrasts with previous schemes, which either they used solely pre-training and therefore were not able to benefit from the representation bias, or they used solely multi-task learning and were not able to tweak the representation bias. 

\begin{figure*}
    \centering
    \begin{subfigure}[b]{0.3\textwidth}
    	\centering
        $p^{const}_q(t) = \alpha$
        \includegraphics[width=\textwidth]{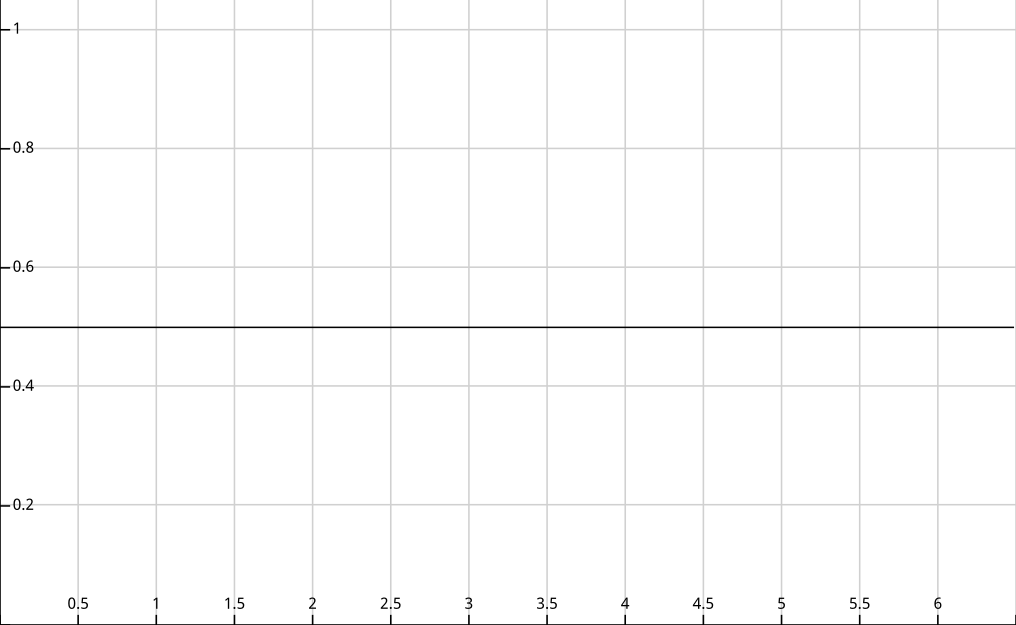}
        \caption{Constant Schedule}
        \label{fig:const}
    \end{subfigure}
    \hfill
    \begin{subfigure}[b]{0.3\textwidth}
		\centering
        $p^{exp}_q(t) = 1 - e^{-\alpha t}$
        \includegraphics[width=\textwidth]{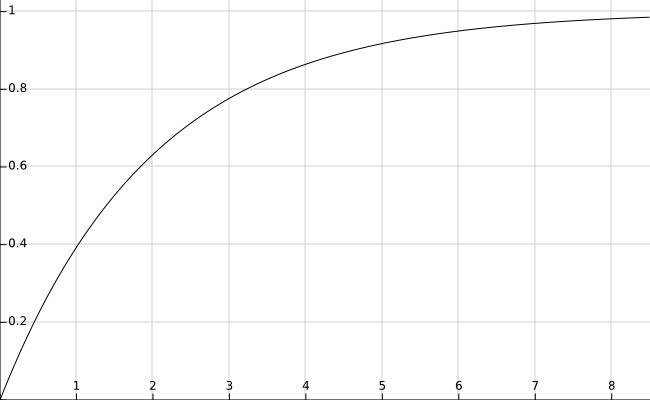} \\
        \caption{Exponential Schedule}
        \label{fig:exp}
    \end{subfigure}
    \hfill
    \begin{subfigure}[b]{0.3\textwidth}
    	\centering
        $p^{sig}_q(t) = \frac{1}{1+e^{-\alpha t}}$
        \includegraphics[width=\textwidth]{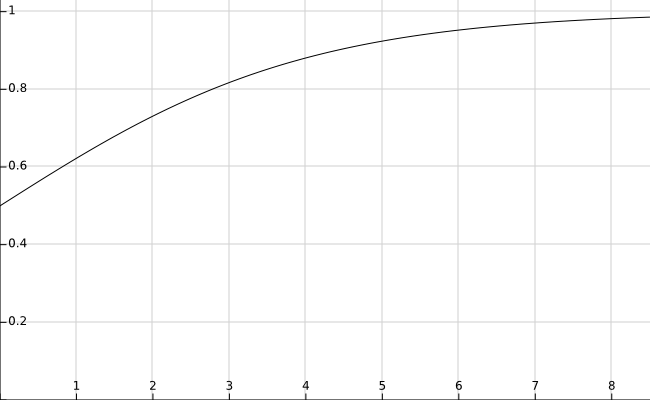}\\
        \caption{Sigmoid Schedule}
        \label{fig:sigmoid}
    \end{subfigure}
    \caption{Illustration of different scheduling strategies determining the probability of the next training example to be picked from each of the multiple tasks we learn. Each sub-plot in the figure matches a different scheduling strategy (with $\alpha$ set to $0.5$). The sub-plot describes the probability ($p$, y-axis) of the task we wish to improve ($q$) using Scheduled Multi-Task Learning as a function of the number of epochs trained ($t$, x-axis) so far. The remaining probability is uniformly distributed among the rest of the tasks.}\label{fig:schedules}
\end{figure*}

We aim to improve generalization over a specific task and dataset (queue) using examples from related tasks and datasets. We suggest three schedulers to do so: \textit{Constant Scheduler}, \textit{Exponential Scheduler}, and \textit{Sigmoid Scheduler} (Figure \ref{fig:schedules}). As input, the schedulers receive the fraction of training epochs done so far (\mbox{$t = \nicefrac{Sent}{\|Corpus\|}$}), and a hyper-parameter ($\alpha$) determining the slope of the scheduler. Given slope parameter ($\alpha$) and the epoch number, the chosen scheduler depicts a multinomial distribution for choosing each of the queues as the source of the next training example.

\paragraph{Constant Scheduler} We assign constant probability to the queue we focus on and divide the rest of the probability uniformly between remaining queues ($p_q(t) = \alpha$). This is similar to previous Multi-Task Learning approaches \cite{DBLP:journals/ml/Caruana97}.

\paragraph{Exponential Scheduler} We assign exponentially increasing probability to the queue we focus on and divide the rest of the probability uniformly between remaining queues ($p_q(t) = 1 - e^{-\alpha t}$). This approach starts by only looking at the training from all the tasks besides the task that we wish to focus on, and it later tunes the parameters based solely on the main task (resembling pre-training and fine-tuning).

\paragraph{Sigmoid Scheduler}  We assign probability to the queue we focus on using a sigmoid and divide the rest of the probability uniformly between remaining queues ($p_q(t) = \frac{1}{1+e^{-\alpha t}}$). This approach starts by looking at all tasks (resembling MTL), and it later tunes the parameters based solely on the main task we wish to focus on.

\section{Experimental Setup}

We evaluate the effectiveness of our models for a low-resource setting and a standard setting. Translation performances are reported in case-sensitive BLEU \cite{DBLP:conf/acl/PapineniRWZ02}. We report translation quality using tokenized\footnote{All texts are tokenized with \texttt{tokenizer.perl} and BLEU scores are computed with \texttt{multi-bleu.perl}.} BLEU comparable with existing Neural Machine Translation papers.

Our experiments are centered around the translation task. We aim to determine whether other syntactically oriented tasks can improve translation and vice versa. Each task is presented in a sequence to sequence manner (as described in Section \ref{sec:MTOSS}). A single sequence to sequence with attention model is used to solve all tasks (all the parameters are shared between the different tasks).

\subsection{Data}
We train the byte-pair encoding model \cite{DBLP:conf/acl/SennrichHB16a} for the translation parallel corpus and apply it to all the data (including non-translation data). 

\paragraph*{Syntax} For English, we extract part-of-speech tagging, dependency heads and labels  from the Penn tree-bank \cite{DBLP:journals/coling/MarcusSM94} with Stanford Dependencies\footnote{Training: 02-21. Development: 22. Test: 23.}. For German, we extract them from TIGER tree-bank \cite{brants2002tiger}.\footnote{German CoNLL 2009 dataset \cite{hajivc-EtAl:2009:CoNLL-2009-ST}.} Both tree-banks are annotated by experts and contain the gold annotations for dependency parsing and part-of-speech tags.

Given the language, we extract three datasets from the relevant tree-bank. We extract parallel corpus of sentences and their gold part-of-speech annotations. The same is done in order to extract a dataset of the unlabeled distances and the dependency labels.

\paragraph*{Translation} In order to simulate low-resource translation tasks, we used 4M tokens of the WIT corpus \cite{cettolo2012wit3} for German to English as training data. We used \emph{tst2012} for validation and \emph{tst2013} for testing, provided by the International Workshop on Spoken Language Translation (IWSLT). Byte-pair encoding is applied, resulting in a vocabulary of 29937 tokens in the source side and 21938 tokens in the target side.

For standard translation setting, we use WMT parallel corpus \cite{buck2014n} with 4.5M sentence pairs (we translate from English to German). We use \emph{newstest2013} (3000 sentences) as the development set to select our hyper-parameters, and \emph{newstest2014} for testing. Note that we use the same (MT) development sets to select the hyper-parameters of the syntactically oriented tasks. After byte-pair encoding is applied, it results in a vocabulary of 59937 tokens in the source side and 63680 tokens in the target side.

We only used training examples shorter than 60 words per sentence. We also filter out pairs where the target length is more than $1.5$x times the source length.

\subsection{Training Details}
We use mini-batching that limits the number of words in the mini-batch instead of the number of sentences \cite{DBLP:journals/corr/MorishitaONYSN17}. We limit the mini-batch size to 5000 words. Based on the scheduler we sample, the dataset to draw training examples from, and add it to the mini-batch until the word limit is reached. \emph{In contrast} to other approaches \cite{DBLP:journals/corr/LuongLSVK15,DBLP:conf/naacl/ZophK16}, our mini-batch is not separated by tasks and often contains examples from multiple tasks.  We shuffle each dataset at the beginning of the training, and after the model has been trained on all the source and target pairs belonging to the dataset(s). 

We use a two layer stacking \textsc{BiLSTM} for the decoder, and a single layer \textsc{BiLSTM} for the encoder. For the low-resource setting, the number of dimensions of the \textsc{LSTM} and the word embedding is set to 250. For the standard setting, the number of dimensions is set to 500. The dimensionality in the standard setting is set to 500 (instead of 1000), in order to enable quick convergence and thereby examine our approach in many different combinations.  The weight updates were determined using the unbiased Adam algorithm \cite{DBLP:journals/corr/KingmaB14}.

We used 0.5 as the scheduler's slope ($\alpha$) (see Section \ref{sec:smtl}) for all our experiments. We use beam search decoding (of size 5) when decoding the test results. For all tasks (including dependency parsing and part-of-speech tagging), we choose the model that maximizes the BLEU score between the reference development corpus and the system prediction on that corpus. For each scheduler and combination of tasks, we report the test score of the model achieving the best development score of three single runs (each with different random initialization).

Our code is implemented in C++, using the DyNet framework \cite{dynet}. When running on a single GPU device Tesla K80, it takes 5-7 days to completely train a model with 4.5 million sentence pairs, and 12 hours for the low resource setup (4M tokens).

\section{Results}
We show the base performance of each task using our Many Tasks One Sequence to Sequence model (subsection \ref{subsec:aux}). We explore multiple combinations of those concurrently learned using Scheduled Multi-Task Learning (subsection \ref{subsec:translation}). We explore (subsection \ref{tunningsch}) different slope parameter ($\alpha$) values (see Section \ref{sec:smtl}) with the intention of optimizing machine translation by leveraging the additional tasks. Finally, we compare our architecture with an architecture that uses separate decoders for each task (Section \ref{comparch}) with a focus on machine translation.

\subsection{Auxiliary Tasks}
\label{subsec:aux}

We use dependency parsing and part-of-speech tagging as auxiliary tasks. Our method utilizes BiLSTM features for syntax as proposed by \newcite{DBLP:journals/tacl/KiperwasserG16} and attention proposed by \newcite{DBLP:journals/corr/DozatM16}, however ours does not impose any tree structure constraints since it is the architecture for translation described in Section \ref{seq2seq}. The model does not even contain the length of the sentence as a hard constraint, meaning that it can arbitrarily output a shorter/longer sequence.\footnote{All evaluation metrics penalize sequences of the wrong length.} Although no structural constraints were imposed, our sequence to sequence model is able to obtain a decent parsing result.\footnote{The parsing only model (without MTL) was trained solely on the unlabeled dependency arcs. Full parsing model that was used in conjunction with other tasks was trained as separate tasks (in an MTL manner) on both unlabeled arcs and their labels.} The model achieves 86.99 UAS for English Penn tree-bank with Stanford Dependencies,\footnote{By increasing the dimensionality of the network for the English parsing task, we achieve results around 90 UAS, but in Table \ref{tab:SmtlWMT} we report results with 500 dimensions since it is the one used in the multi-task learning scenario with the WMT data (see Section \ref{subsec:translation}).} and 80.28 UAS for the German TIGER tree-bank when the model is only trained to predict the sequence of distances to head as described in Section \ref{sec:MTOSS}. This is below the best results achieved by state-of-the-art parsers, that are already around 95 for English \cite{DBLP:journals/corr/DozatM16,kuncoro-EtAl:2017:EACLlong}, and around 90 for the same German dataset \cite{andor-EtAl:2016:P16-1,kuncoro-EtAl:2016:EMNLP2016,bohnet-nivre:2012:EMNLP-CoNLL}. As a side product of our research, we show that dependency parsing can be approached via a sequence to sequence with an attention mode commonly used for neural machine translation with linearized (using sequences of head distances) dependency trees. Note that, in this case, the models are solely trained on predicting the sequence of distances to the head and are not trained to predict the sequence of dependency labels.

For part-of-speech tagging, we use the same sequence to sequence with attention architecture presented in Section \ref{seq2seq}.  Our model uses BiLSTM encodings, in a similar way as  proposed by \newcite{DBLP:journals/corr/WangQSHZ15} for part-of-speech tagging. Similarly as in parsing (see above), we do not force one part-of-speech per word and do not force the model to scan the sentence linearly nor do we add any hard constraints on the length. Even without these constraints, the model achieves accuracy of 95.07 for English Penn tree-bank and 95.41 for German TIGER treebank, which is lower than the best systems that achieve results above 97 \cite{andor-EtAl:2016:P16-1,bohnet-nivre:2012:EMNLP-CoNLL} for both languages. We use the same datasets as in the parsing task. 

Note that both for part-of-speech tagging and dependency parsing, our models are trained with byte-pair encoding (BPE) in the input side \cite{DBLP:journals/corr/SennrichHB15}, meaning that there are usually more tokens in the input than in the output (which has exactly one label or a token representing the distance to head per word). For the single-task models we also use 250 dimensions for our network (word embeddings, hidden dimensions and LSTM input dimensions) for German and 500 dimensions for English.

\subsection{Translation Task}
\label{subsec:translation}
We start from our baseline system which achieves results which are comparable (see Tables \ref{tab:SmtlIWSLT} and \ref{tab:SmtlWMT}) to the ones reported by  \newcite{DBLP:journals/corr/BahdanauCB14} on the standard setting (WMT), and \newcite{DBLP:conf/wmt/NiehuesC17} on the low-resource setting (IWSLT). We examine the effect of Scheduled Multi-Task Learning on the translation quality compared to the baseline system with a constant value of the slope parameter ($\alpha$) set to 0.5.\footnote{In Section \ref{tunningsch} we explore different slope parameter ($\alpha$) values for the same task.} We also show that amount of representation bias the models chose to obtain by testing each model on each of the auxiliary tasks.

As in part-of-speech tagging and dependency parsing (both predicting a sequence of heads and dependency labels, as separate tasks. This is the reason why we report LAS), we use BPE encoding both in target and source. We use 250 dimensions for the low-resource setting (IWSLT) and 500 dimensions for the standard setting (WMT).

\paragraph*{Low-Resource Setting}
In a low-resource setting, we witness a significant increase in translation quality when doing basic multi-task learning (with the constant scheduler) with syntactic auxiliary tasks (Table \ref{tab:SmtlIWSLT}). We attribute this to the additional linguistic information which is difficult to learn from a low-resource setting. The latter can be observed in Table \ref{tab:SmtlIWSLT} which shows an increase of roughly 2.7 BLEU when adding part-of-speech information and 1.85 BLEU when adding dependency parsing.

The baseline (constant) multi-task learning scheduler reaches better translation quality than the sigmoid and exponential scheduler. We hypothesize that in a low-resource setting a strong representation bias incorporating linguistic knowledge helps to build generalized representation which cannot be obtained from a relatively small parallel corpus.

We evaluate the dependency parsing scores and the part-of-speech tagging accuracy of the models tuned to perform translation on the held-out development set. The percentage of correctly predicted unlabeled arcs by MTL is no more than 10 UAS points worse compared to the models that are solely train to parse or to tag, and  they are very close for the Constant Scheduler. Note that the models are optimized to perform translation, however they are still able to parse sentences with a reasonable accuracy. MTL models are also better at translation than models trained on the vanilla translation data. This means that the attentional model of translation is benefiting from the syntactic information, and therefore chooses to learn parameters close to the syntactically oriented tasks, even though there are no constraints forcing it to do so.

\begin{table*}[h!]
\centering
\scalebox{0.77}{
\begin{tabular}{l |c |c |c |c |c}
Scheduler & Tasks & BLEU & POS & UAS & LAS \\
\hline \hline
\multirow{3}{*}{No MTL} & NMT & 27.70 & -- & -- & -- \\
 & POS & -- & 95.41 & -- & --\\
 & Parsing (Unlabeled) & -- & -- & 80.28 & -- \\
\hline \hline
\multirow{3}{*}{Constant Scheduler} & NMT + POS & \textbf{30.4} & 93.51 & -- & -- \\
& NMT + Parsing & 28.73 & -- & 79.78 & 74.25 \\
& NMT + POS + Parsing &  29.08 & 94.80 & 79.38 & 74.13\\
\hline \hline
\multirow{3}{*}{Exponent Scheduler} & NMT + POS & 30.15 & 89.05 & -- & -- \\
& NMT + Parsing & 29.37 & -- & 67.60 & 60.71 \\
& NMT + POS + Parsing & 29.55 & 91.48 & 72.85 & 66.44\\
\hline \hline
\multirow{3}{*}{Sigmoid Scheduler} & NMT + POS & 30.2 & 90.74 & -- & -- \\
& NMT + Parsing & 28.78 & -- & 69.26 & 62.43 \\
& NMT + POS + Parsing & 28.93 & 89.11 & 65.92 & 58.46\\
\end{tabular}
}
  \caption{Scheduled Multi-Task learning results for IWSLT German to English translation.} \label{tab:SmtlIWSLT}
\end{table*}

As mentioned above, the automatic scores show a significant improvement over the NMT system that only sees the parallel sentences. In Table \ref{tab:example1}, we show some randomly picked examples from the IWSLT development data in order to show how each of the systems performed. We include Google web\footnote{\url{https://translate.google.com}} system to see a comparison with a state-of-the-art system that is probably trained with more data. Note that in the low-resource data we only have ~300k sentence pairs. We selected the output of the systems with highest score in each category (NMT Only, NMT+POS with Constant Scheduler, NMT+Parsing and NMT+POS+Parsing with Exponent Scheduler). 

\begin{table*}[h!]
\centering
\scalebox{0.67}{
\begin{tabular}{|l |l|}
\hline 
System & Example \\
\hline \hline
Source & Jeden Tag nahmen wir einen anderen Weg , sodass niemand erraten konnte , wohin wir gingen .\\
\hline
Google & Every day we took a different route so no one could guess where we were going.\\
NMT Only & We took another way for us to guess that no one could guess where we left.\\
NMT+POS  & Every day we took another way so no one could guess where we went. \\
NMT+Parsing & Every day we took another way that no one could guess where we went. \\
NMT+POS+Parsing & Every day we took another way that no one could guess where we were going.\\
\hline \hline
Source & Wissen Sie, wie viele Entscheidungen Sie an einem typischen Tag machen ?\\
\hline
Google & Do you know how many decisions you make on a typical day?
\\
NMT Only & You know how many decisions you make on a typical day?\\
NMT+POS  & You know how many decisions you make on a typical day? \\
NMT+Parsing & Do you know how many decisions you make on a typical day? \\
NMT+POS+Parsing & Do you know how many decisions you make on a typical day?\\
\hline \hline
Source & Im Winter war es gem\"utlich, aber im Sommer war es unglaublich hei{\ss}.\\
\hline
Google & in winter it was cozy, but in the summer it was incredibly hot.\\
NMT Only & In winter, it was comfortable, but it was incredibly hot.\\
NMT+POS  & In winter, it was comfortable, but in summer it was incredibly hot.\\
NMT+Parsing & In the winter, it was comfortable, but in the summer it was incredibly hot. \\
NMT+POS+Parsing & In the winter, it was comfortable, but in summer it was incredibly hot.\\
\hline
\end{tabular}
}
\caption{Examples from our low-resource IWSLT German to English translation.} \label{tab:example1}
\end{table*}

Given that the examples in Table \ref{tab:example1} suggest that the SMTL models may be doing a better job at avoiding
dropping words we complement our BLEU scores with the METEOR evaluation metric \cite{Lavie:2007:MAM:1626355.1626389} which is more sensitive to recall. We report METEOR (and fragmentation penalty that captures how well the system produces the correct order of the words) for the models with highest BLEU scores in each category (NMT Only, NMT+POS with Constant Scheduler, NMT+Parsing and NMT+POS+Parsing with Exponent Scheduler). Table \ref{tab:meteor} shows the results. Models with the higher BLEU scores also produce higher METEOR scores. In addition it is interesting to see that the fragmentation penalty is higher for the NMT Only model; the NMT Only model only produces 19,768 test words (for the entire test set) while the rest produce longer sentences with more than 20,400 test words. All of this suggests that the additional tasks are helping to avoid dropping parts of the sentence which leads to more adequate outputs.

\begin{table}[h!]
\centering
\scalebox{0.7325}{
\begin{tabular}{|l|c|c|c|}
\hline 
System & BLEU & Fragmentation & METEOR \\
\hline \hline
\hline
NMT Only & 27.7 & 50.36 & 30.91 \\
NMT+POS  & \textbf{30.4} & \textbf{49.95} & \textbf{31.83} \\
NMT+Parsing & 29.37 & 50.14 & 31.43 \\
NMT+POS+Parsing & 29.55 & 50.02 & 31.56 \\
\hline
\end{tabular}
}
\caption{METEOR results for our best scoring systems in comparison with BLEU scores. Fragmentation refers to the fragmentation penalty.} \label{tab:meteor}
\end{table}

\paragraph*{Standard Setting}
In the standard-resource setting (Table \ref{tab:SmtlWMT}), the exponent scheduler (when using part-of-speech tagging as an auxiliary task) achieves significantly better numbers than the other multi-task learning strategies, and achieves a translation quality that surpasses the base neural translation system (by 0.7 BLEU points). When applying the Constant Scheduler (basic multi-task learning) we see a deduction of at least 1 BLEU point compared to the score of the translation without multi-task learning. We assume that additional out-of-domain linguistic knowledge (such as syntax in the Penn tree-bank) might confuse the linguistic properties that the translation model is inferring from the comparably large machine translation data.

The sigmoid scheduler reaches better translation quality than the constant scheduler by roughly 1 BLEU point (and improves over the base neural translation system) and it improves over the Exponent Scheduler for the tasks that include the parsing objective. This suggests that putting more emphasis on syntax regularizes the model towards capturing linguistic properties (as exponential scheduler does), but that focusing on them as the training continues causes a representation bias which puts focus on out-of-domain data, which, as a result, degrades the translation quality.

Similarly to the low-resource setting, we evaluate the dependency parsing scores and the part-of-speech tagging accuracy of the models tuned to perform translation on the held-out development set. The result for the standard setting shows a drop (of 12 UAS point at most) in the parsing accuracy when trained in a multi-task manner. The accuracy of the part-of-speech tagger improves when using constant and sigmoid schedulers. The part-of-speech accuracy plunges significantly when using the exponential scheduler; and in turn, the translation quality raises by 0.7 BLEU over the baseline model. This suggests that softening the representation bias (by allowing the model to gradually fine-tune on translation) is necessary to improve the translation task. When adding dependency parsing and part-of-speech tagging, we do not see a significant drop in those auxiliary tasks and also the results for translation does not improve. This might suggest that representation bias is too strict in this case and does not allow the representation to learn beyond the auxiliary tasks.  

\begin{table*}[h!]
\centering
\scalebox{0.77}{
\begin{tabular}{l |c |c |c |c |c}
Scheduler & Tasks & BLEU & POS & UAS & LAS \\
\hline \hline
\multirow{3}{*}{No MTL} & NMT & 19.30 & -- & -- & -- \\
 & POS & -- & 95.07 & -- & --\\
 & Parsing (Unlabeled) & -- & -- & 86.99 & -- \\
\hline \hline
\multirow{3}{*}{Constant Scheduler} & NMT + POS & 18.29 & 95.73 & -- & -- \\
& NMT + Parsing & 17.87 & -- & 85.74 & 81.65 \\
& NMT + POS + Parsing & 18.09 & 96.30 & 86.58 & 82.83\\
\hline \hline
\multirow{3}{*}{Exponent Scheduler} & NMT + POS & \textbf{20.02} & 89.89 & -- & -- \\
& NMT + Parsing & 18.85 & -- & 80.40 & 74.70 \\
& NMT + POS + Parsing & 18.04 & 94.68 & 82.54 & 77.57\\
\hline \hline
\multirow{3}{*}{Sigmoid Scheduler} & NMT + POS & 19.21 & 95.20 & -- & -- \\
& NMT + Parsing & 19.08 & -- & 75.42 & 69.27\\
& NMT + POS + Parsing & 19.26 & 94.66 & 80.33 & 75.15\\
\end{tabular}
}
  \caption{Scheduled Multi-Task learning results for WMT14 English to German translation.} \label{tab:SmtlWMT}
\end{table*}

In order to complement our automatic scores, we performed simple human evaluation, in which an independent German native speaker (who is also proficient in English) scored 50 sentences from 0 to 5 (being 0 exceptionally poor, and 5 excellent); the sentences were randomly shuffled so there is no bias towards the position in which they were presented. The NMT only system achieved a score of 2.54, the best system with part-of-speech tagging only (which is the constant scheduler) achieved 2.68, and both systems that incorporate dependency parsing (NMT+Parsing and NMT+POS+Parsing with the sigmoid scheduler) achieve 2.78 in average. An example output of the systems, also compared to Google, is shown in Table \ref{tab:example2}; we observe how the system that uses all auxiliary tasks manages to get the gender agreement right for the words \emph{journalist} and \emph{Katie}.

\begin{table*}[h!]
\centering
\scalebox{0.65}{
\begin{tabular}{|l|l|}
\hline
System & Example \\
\hline \hline
Source & In an interview with US journalist Katie Couric , which is to be broadcast on Friday ( local time ) , Bloom said , \\ &" sometimes life does n't go exactly as we plan or hope for " .\\
\hline
Google & In einem Interview mit der US-Journalistin Katie Couric, das am Freitag (Ortszeit) ausgestrahlt wird, sagte Bloom: \\ & "Manchmal l\"auft das Leben nicht genau so, wie wir es planen oder erhoffen".\\
NMT Only & In einem Interview mit der US - Journalist Katie Couric, das am Freitag (Ortszeit) verbreitet werden soll, sagte Bloom, \\ &"manchmal geht das Leben nicht genau wie wir planen oder Hoffnung f\"ur".\\
NMT+POS  & In einem Interview mit den US - Journalisten Katie Couric, die am Freitag (Ortszeit) ausgestrahlt werden soll, sagte Bloom: \\ &"Manchmal ist das Leben nicht genau so, wie wir es planen oder hoffen." \\
NMT+Parsing & In einem Interview mit dem US - Journalist Katie Couric, der am Freitag gesendet wird (Ortszeit) , sagte Bloom, \\ &"manchmal wird das Leben nicht genau so aussehen, wie wir uns vorstellen oder hoffen". \\
NMT+POS+Parsing & In einem Interview mit US - Journalistin Katie Couric , das am Freitag ausgestrahlt wird (Ortszeit), sagte Bloom: \\ &"Manchmal ist das Leben nicht genau so, wie wir planen oder hoffen".\\
\hline
\end{tabular}
}
  \caption{Example from our standard English to German translation (WMT).} \label{tab:example2}
\end{table*}

\begin{figure*}[ht]
    \centering
    \begin{subfigure}[b]{0.3\textwidth}
    	\centering
        \scalebox{0.73}{
\begin{tikzpicture}
\begin{axis}[
    title=Constant Scheduler,
    xlabel={$\alpha$},
    ylabel={BLEU Score},
    grid=major,
    ymin=27.8,
    ymax=29.8
]
\addplot+[red,line width=0.3mm, mark=square*] table {sch1_pos.dat};
\addplot+[blue,line width=0.3mm, mark=star] table {sch1_parsing.dat};
\addplot+[teal,line width=0.3mm, mark=o] table {sch1_parsing_pos.dat};
\end{axis}
\end{tikzpicture}}
\end{subfigure}
\hfill
    \begin{subfigure}[b]{0.3\textwidth}
    	\centering
        \scalebox{0.73}{
\begin{tikzpicture}
\begin{axis}[
    title=Exponential Scheduler,
    xlabel={$\alpha$},
    ylabel={},
    grid=major,
    ymin=27.8,
    ymax=29.8
]
\addplot+[red,line width=0.3mm, mark=square*] table {sch2_pos.dat};
\addplot+[blue,line width=0.3mm, mark=star] table {sch2_parsing.dat};
\addplot+[teal,line width=0.3mm, mark=o] table {sch2_parsing_pos.dat};
\end{axis}
\end{tikzpicture}}
\end{subfigure}
\hfill
    \begin{subfigure}[b]{0.3\textwidth}
    	\centering
        \scalebox{0.73}{
        \begin{tikzpicture}
\begin{axis}[
    title=Sigmoid Scheduler,
    xlabel={$\alpha$},
    ylabel={},
    grid=major,
    ymin=27.8,
    ymax=29.8
]
\addplot+[red,line width=0.3mm, mark=square*] table {sch3_pos.dat};
\addplot+[blue,line width=0.3mm, mark=star] table {sch3_parsing.dat};
\addplot+[teal,line width=0.3mm, mark=o] table {sch3_parsing_pos.dat};
\end{axis}
\end{tikzpicture}}
\end{subfigure}
\caption{A plot of the BLEU score for the three schedulers over different alpha values. The BLEU score is the average test score of four independent experiments trained on IWSLT training set optimized for maximal score on the IWSLT development set. We use $0 \leq \alpha < 1$ for the constant scheduler, and $0 \leq \alpha \leq 3$ for both the exponential and the sigmoid schedulers. The red line (squares) represents result with part-of-speech tagging as auxiliary task; blue line (asterisks) represents result with parsing as auxiliary task; teal line (circles) represents the results with both tasks as auxiliary tasks. \label{fig:varyingw}}
\end{figure*}
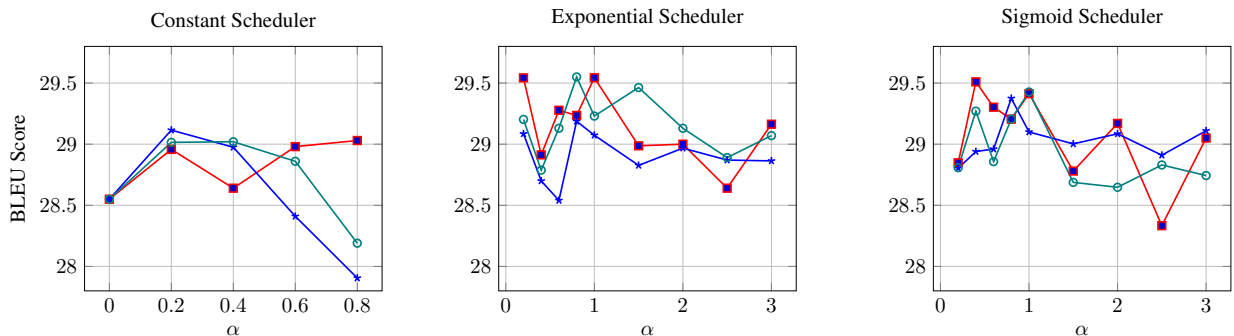

\subsection{Scheduler Tuning}
\label{tunningsch}
We study the impact of different slope parameter ($\alpha$) values on the translation BLEU score using the low-resource IWSLT corpus. For each scheduler, we train the model (pick the model performing best on the development set) four times with multiple $\alpha$ values and different auxiliary tasks, and average the BLEU score of the decoded test set (Figure \ref{fig:varyingw}).

 We compare the average result of the Constant Scheduler (Figure \ref{fig:varyingw}) against the result of the best performing model on the development set (Table \ref{tab:SmtlIWSLT}). The average result when training with auxiliary tasks (i.e. the Constant Scheduler where $\alpha$ is set to zero) is significantly higher than the result of the best model on the development set (0.7 BLEU points), the matching scores are 28.5 and 27.7 BLEU points. The average score when using the Constant Scheduler with $\alpha$ set to half is greater than the score of the best performing model on the development set. The average result of the constant scheduler setting suggests that multi-task learning helps to mitigate over-fitting.

The average results of a model with both parsing and part-of-speech tagging peak when the slope parameter ($\alpha$) is approximately 1 for both the exponential scheduler (29.43 BLEU) and the sigmoid scheduler (29.55 BLEU). For those schedulers, if the $\alpha$ value is high, the probability of training on the auxiliary tasks decreases more rapidly. This suggests that the model needs syntactically oriented synergistic tasks to guide the initial steps to improve convergence; after four epochs the probability of training on an auxiliary task is negligible. The constant scheduler peaks when alpha is $0.2$ (yielding an average score of 29.03 BLEU), suggesting that enriching the representation with a small amount of syntactical information helps. This confirms our intuition that syntax is helpful.

Looking at the constant scheduler, which performed best for this dataset (Table \ref{tab:SmtlIWSLT}), we see that the best result is achieved by using parsing as the single auxiliary task (without parts-of-speech). This hints that parsing has potential to help machine translation, even more than part-of-speech tagging with constant scheduler \cite{DBLP:conf/wmt/NiehuesC17}.

\subsection{Architecture Comparison}
\label{comparch}

In order to further validate that the contribution of Scheduled Multi-Task Learning is not limited to our chosen sequence to sequence architecture, we study the impact of our method with a single (and shared across tasks) encoder and the architecture of separate decoders which has already proven to be a very effective multi-task learning scheme \cite{DBLP:journals/corr/LuongLSVK15,DBLP:conf/wmt/NiehuesC17}. In the latter, each of the decoders is responsible for a different task (i.e. syntax, parts-of-speech, translation, etc.) using a single representation generated by the shared encoder.

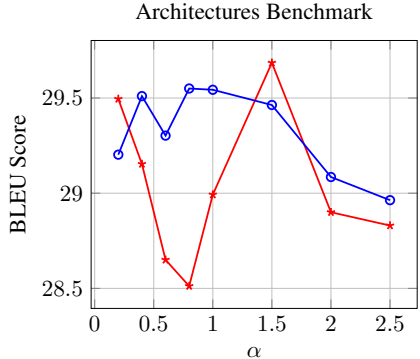
\begin{figure}[h]
\centering
\scalebox{0.8}{
\begin{tikzpicture}
\begin{axis}[
    title=Architectures Benchmark,
    xlabel={$\alpha$},
    ylabel={BLEU Score},
    grid=major,
]
\addplot+[red,line width=0.3mm, mark=star] table {maxavgshareddec.dat};
\addplot+[blue,line width=0.3mm, mark=o] table {maxavgsharedall.dat};
\end{axis}
\end{tikzpicture}}
\caption{A plot of the best BLEU score for each alpha value using our approach (blue line with circle) and separate decoders (red line with asterisks). The BLEU score is the average test score of four independent experiments for each scheduler and each value of alpha trained on IWSLT training set optimized for maximal score on the development set.}
\label{fig:diffarc}
\end{figure}

In Figure \ref{fig:diffarc}, we show the comparison between our Many Tasks One Sequence to Sequence architecture (Section \ref{sec:MTOSS}) and the architecture of separate decoders by using the IWSLT data set. We report BLEU scores as the average test score of four independent experiments for each scheduler and each value of the slope parameter $\alpha$. The plot shows the average for all schedulers. The best average score for most of the alpha values is greater than the average score without Scheduled Multi-Task Learning (28.5 BLEU). We conclude that scheduled multi-task with syntactic auxiliary tasks is helpful not solely for our architecture, but potentially for other systems as well.

The architecture of separate decoders and a shared encoder peaks at 29.68 BLEU which is higher than the peak score of the shared decoder architecture (29.55 BLEU) by 0.15 BLEU points. The best result of the separate decoders significantly varies as alpha is changed ($\sigma=0.38$). The result of the shared decoder architecture also varies for different alpha, but in a more subtle manner ($\sigma=0.21$). This suggests that the separate decoders architecture is more sensitive to the scheduler used than the shared decoders architecture.  

\section{Discussion}

Scheduled Multi-Task Learning is complementary to other transfer learning methods like pre-training and fine-tuning. It is common to use pre-training in the form of word embeddings \cite{NIPS2013_5021,goldberg2017neural}. One advantage of pre-trained word embeddings is the representation of out-of-vocabulary (OOV) words. Through pre-training, OOV words are commonly trained using an early-stopping methodology so their representation remains close to words in the training corpus, thus enabling the model to generalize for unseen words and achieve higher performance in the final task. This constraint limits the flexibility of the optimizer to choose better word representation for words within the training corpus. Scheduled multi-task learning (and the exponential scheduler in particular) mitigates this problem by allowing the representation of the final task and the auxiliary tasks to be tuned to best fit each other.

The exponential scheduler starts by pre-training the model on an auxiliary task (in our case, part-of-speech tagging and dependency parsing) and gradually puts more focus on our main task (NMT). This enables the model to start with a representation which is able to solve structured prediction tasks containing linguistic knowledge; as the training progresses and the focus is shifted by the scheduler towards the main task, the OOV words representations continue to represent the syntax objective since the auxiliary tasks are less visited but still in use during training. Having embeddings that share the same space enables the model to share information between the tasks, and functions as regularization \cite{DBLP:journals/ml/Caruana97}. The effectiveness of this scheduler is supported by the results (Table \ref{tab:SmtlIWSLT}) showing superior results (on average) on the WIT German to English translation task.

Many approaches have been employing multi-task learning in order to inject linguistic knowledge with great success \cite[inter-alia]{DBLP:conf/emnlp/LuongPM15,DBLP:conf/wmt/NiehuesC17,martinezalonso-plank:2017:EACLlong}. The final representation is then adapted to solve multiple tasks, however continuing to fine-tune on solely the main task might result in better accuracy. The latter resembles the Sigmoid Scheduler which starts with multi-task learning and gradually shifts to fine-tuning. The results (Table \ref{tab:SmtlWMT}) support that this approach can further benefit multi-task learning systems since it shows superior results (on average) in the WMT14 English to German translation task, although it is still not more superior than the baseline that does not use MTL.





\section{Conclusions and Future Work}

This paper presents an architecture to perform multi-task learning focusing on the attentional model of translation jointly with linearized dependency parsing and part-of-speech tagging. We show how diverse scheduling strategies perform differently and help to improve the scores in a low-resource setting and a standard setting (bigger dataset). 
The exponent scheduler achieves the best results on average and the trained models still remember how to perform the auxiliary tasks (part-of-speech tagging and dependency parsing). 
This means that a key aspect of our models is that they are able to improve the translation accuracy by incorporating syntactically based objectives into the model. Our models report modest dependency parsing and part-of-speech tagging numbers but they clearly learn to perform the tasks; it is worth noting that there is a lack of constraints related to sequence length and correspondence between input tokens and tags/distances which is needed to achieve good parsing scores \cite{zhang-EtAl:2017:EMNLP20173}.

We also want to explore another family of schedulers which treats the layers of the neural network differently. For instance, the scheduler can gradually freeze the top LSTM layer of the decoder (by lowering the learning rate), allowing fine-tuning only of the bottom LSTM layer when training for auxiliary tasks. \newcite{sogaard2016deep} demonstrated the potential of such an approach. Our experiments show that scheduled multi-task learning is very sensitive to the type of scheduler chosen, and many types of schedulers can be explored. We plan to carry out these experiments in the future.

\section*{Acknowledgments}
Many thanks to Todd Ward, Wael Hamza, Yaser Al-Onaizan, Yoav Goldberg, the three anonymous reviewers and the action editor for their useful comments that improved the final version of this paper.

\bibliography{tacl}

\begin{thebibliography}{}

\bibitem[\protect\citename{Abu{-}Mostafa}1990]{DBLP:journals/jc/Abu-Mostafa90}
Yaser~S. Abu{-}Mostafa.
\newblock 1990.
\newblock Learning from hints in neural networks.
\newblock {\em J. Complexity}, 6(2):192--198.

\bibitem[\protect\citename{Aharoni and
  Goldberg}2017]{DBLP:conf/acl/AharoniG17a}
Roee Aharoni and Yoav Goldberg.
\newblock 2017.
\newblock Towards string-to-tree neural machine translation.
\newblock In {\em Proceedings of the 55th Annual Meeting of the Association for
  Computational Linguistics, {ACL} 2017, Vancouver, Canada, July 30 - August 4,
  Volume 2: Short Papers}, pages 132--140.

\bibitem[\protect\citename{Ammar \bgroup et al.\egroup
  }2016]{DBLP:journals/tacl/AmmarMBDS16}
Waleed Ammar, George Mulcaire, Miguel Ballesteros, Chris Dyer, and Noah~A.
  Smith.
\newblock 2016.
\newblock Many languages, one parser.
\newblock {\em {TACL}}, 4:431--444.

\bibitem[\protect\citename{Andor \bgroup et al.\egroup
  }2016]{andor-EtAl:2016:P16-1}
Daniel Andor, Chris Alberti, David Weiss, Aliaksei Severyn, Alessandro Presta,
  Kuzman Ganchev, Slav Petrov, and Michael Collins.
\newblock 2016.
\newblock Globally normalized transition-based neural networks.
\newblock In {\em Proceedings of the 54th Annual Meeting of the Association for
  Computational Linguistics (Volume 1: Long Papers)}, pages 2442--2452, Berlin,
  Germany, August. Association for Computational Linguistics.

\bibitem[\protect\citename{Bahdanau \bgroup et al.\egroup
  }2014]{DBLP:journals/corr/BahdanauCB14}
Dzmitry Bahdanau, Kyunghyun Cho, and Yoshua Bengio.
\newblock 2014.
\newblock Neural machine translation by jointly learning to align and
  translate.
\newblock {\em CoRR}, abs/1409.0473.

\bibitem[\protect\citename{Bastings \bgroup et al.\egroup
  }2017]{DBLP:conf/emnlp/BastingsTAMS17}
Joost Bastings, Ivan Titov, Wilker Aziz, Diego Marcheggiani, and Khalil
  Sima'an.
\newblock 2017.
\newblock Graph convolutional encoders for syntax-aware neural machine
  translation.
\newblock In {\em Proceedings of the 2017 Conference on Empirical Methods in
  Natural Language Processing, {EMNLP} 2017, Copenhagen, Denmark, September
  9-11, 2017}, pages 1957--1967.

\bibitem[\protect\citename{Baxter}2000]{DBLP:journals/jair/Baxter00}
Jonathan Baxter.
\newblock 2000.
\newblock A model of inductive bias learning.
\newblock {\em JAIR}, 12:149--198.

\bibitem[\protect\citename{Belinkov \bgroup et al.\egroup
  }2017]{belinkov-EtAl:2017:Long}
Yonatan Belinkov, Nadir Durrani, Fahim Dalvi, Hassan Sajjad, and James Glass.
\newblock 2017.
\newblock What do neural machine translation models learn about morphology?
\newblock In {\em Proceedings of the 55th Annual Meeting of the Association for
  Computational Linguistics (Volume 1: Long Papers)}, pages 861--872,
  Vancouver, Canada, July. Association for Computational Linguistics.

\bibitem[\protect\citename{Bingel and
  S{\o}gaard}2017]{bingel-sogaard:2017:EACLshort}
Joachim Bingel and Anders S{\o}gaard.
\newblock 2017.
\newblock Identifying beneficial task relations for multi-task learning in deep
  neural networks.
\newblock In {\em Proceedings of the 15th Conference of the European Chapter of
  the Association for Computational Linguistics: Volume 2, Short Papers}, pages
  164--169, Valencia, Spain, April. Association for Computational Linguistics.

\bibitem[\protect\citename{Bohnet and
  Nivre}2012]{bohnet-nivre:2012:EMNLP-CoNLL}
Bernd Bohnet and Joakim Nivre.
\newblock 2012.
\newblock A transition-based system for joint part-of-speech tagging and
  labeled non-projective dependency parsing.
\newblock In {\em Proceedings of the 2012 Joint Conference on Empirical Methods
  in Natural Language Processing and Computational Natural Language Learning},
  pages 1455--1465, Jeju Island, Korea, July. Association for Computational
  Linguistics.

\bibitem[\protect\citename{Bojar \bgroup et al.\egroup
  }2016]{DBLP:conf/wmt/BojarCFGHHJKLMN16}
Ondrej Bojar, Rajen Chatterjee, Christian Federmann, Yvette Graham, Barry
  Haddow, Matthias Huck, Antonio Jimeno{-}Yepes, Philipp Koehn, Varvara
  Logacheva, Christof Monz, Matteo Negri, Aur{\'{e}}lie N{\'{e}}v{\'{e}}ol,
  Mariana~L. Neves, Martin Popel, Matt Post, Raphael Rubino, Carolina Scarton,
  Lucia Specia, Marco Turchi, Karin~M. Verspoor, and Marcos Zampieri.
\newblock 2016.
\newblock Findings of the 2016 conference on machine translation.
\newblock In {\em Proceedings of the First Conference on Machine Translation,
  {WMT} 2016, colocated with {ACL} 2016, August 11-12, Berlin, Germany}, pages
  131--198.

\bibitem[\protect\citename{Brants \bgroup et al.\egroup }2002]{brants2002tiger}
Sabine Brants, Stefanie Dipper, Silvia Hansen, Wolfgang Lezius, and George
  Smith.
\newblock 2002.
\newblock The {TIGER} treebank.
\newblock In {\em Proceedings of the Workshop on Treebanks and Linguistic
  Theories}, volume 168.

\bibitem[\protect\citename{Buck \bgroup et al.\egroup }2014]{buck2014n}
Christian Buck, Kenneth Heafield, and Bas Van~Ooyen.
\newblock 2014.
\newblock N-gram counts and language models from the common crawl.
\newblock In {\em LREC}, volume~2, page~4.

\bibitem[\protect\citename{Caruana}1997]{DBLP:journals/ml/Caruana97}
Rich Caruana.
\newblock 1997.
\newblock Multitask learning.
\newblock {\em Machine Learning}, 28(1):41--75.

\bibitem[\protect\citename{Cettolo \bgroup et al.\egroup
  }2012]{cettolo2012wit3}
Mauro Cettolo, Christian Girardi, and Marcello Federico.
\newblock 2012.
\newblock Wit3: Web inventory of transcribed and translated talks.
\newblock In {\em Proceedings of the 16th Conference of the European
  Association for Machine Translation (EAMT)}, volume 261, page 268.

\bibitem[\protect\citename{Collobert \bgroup et al.\egroup
  }2011]{DBLP:journals/jmlr/CollobertWBKKK11}
Ronan Collobert, Jason Weston, L{\'{e}}on Bottou, Michael Karlen, Koray
  Kavukcuoglu, and Pavel~P. Kuksa.
\newblock 2011.
\newblock Natural language processing (almost) from scratch.
\newblock {\em Journal of Machine Learning Research}, 12:2493--2537.

\bibitem[\protect\citename{Dozat and Manning}2016]{DBLP:journals/corr/DozatM16}
Timothy Dozat and Christopher~D. Manning.
\newblock 2016.
\newblock Deep biaffine attention for neural dependency parsing.
\newblock {\em CoRR}, abs/1611.01734.

\bibitem[\protect\citename{Dyer \bgroup et al.\egroup
  }2016]{dyer-EtAl:2016:N16-1}
Chris Dyer, Adhiguna Kuncoro, Miguel Ballesteros, and Noah~A. Smith.
\newblock 2016.
\newblock Recurrent neural network grammars.
\newblock In {\em Proceedings of the 2016 Conference of the North American
  Chapter of the Association for Computational Linguistics: Human Language
  Technologies}, pages 199--209, San Diego, California, June. Association for
  Computational Linguistics.

\bibitem[\protect\citename{Dyer}2017]{dyer:2017:CoNLL}
Chris Dyer.
\newblock 2017.
\newblock Should neural network architecture reflect linguistic structure?
\newblock In {\em Proceedings of the 21st Conference on Computational Natural
  Language Learning (CoNLL 2017)}, page~1, Vancouver, Canada, August.
  Association for Computational Linguistics.

\bibitem[\protect\citename{Erhan \bgroup et al.\egroup
  }2010]{DBLP:journals/jmlr/ErhanBCMVB10}
Dumitru Erhan, Yoshua Bengio, Aaron~C. Courville, Pierre{-}Antoine Manzagol,
  Pascal Vincent, and Samy Bengio.
\newblock 2010.
\newblock Why does unsupervised pre-training help deep learning?
\newblock {\em Journal of Machine Learning Research}, 11:625--660.

\bibitem[\protect\citename{Eriguchi \bgroup et al.\egroup
  }2017]{eriguchi-tsuruoka-cho:2017:Short}
Akiko Eriguchi, Yoshimasa Tsuruoka, and Kyunghyun Cho.
\newblock 2017.
\newblock Learning to parse and translate improves neural machine translation.
\newblock In {\em Proceedings of the 55th Annual Meeting of the Association for
  Computational Linguistics (Volume 2: Short Papers)}, pages 72--78, Vancouver,
  Canada, July. Association for Computational Linguistics.

\bibitem[\protect\citename{Goldberg}2017]{goldberg2017neural}
Yoav Goldberg.
\newblock 2017.
\newblock Neural network methods for natural language processing.
\newblock {\em Synthesis Lectures on Human Language Technologies},
  10(1):1--309.

\bibitem[\protect\citename{Graves}2012]{DBLP:series/sci/2012-385}
Alex Graves.
\newblock 2012.
\newblock {\em Supervised Sequence Labelling with Recurrent Neural Networks},
  volume 385 of {\em Studies in Computational Intelligence}.
\newblock Springer.

\bibitem[\protect\citename{Haji\v{c} \bgroup et al.\egroup
  }2009]{hajivc-EtAl:2009:CoNLL-2009-ST}
Jan Haji\v{c}, Massimiliano Ciaramita, Richard Johansson, Daisuke Kawahara,
  Maria~Ant\`{o}nia Mart\'{\i}, Llu\'{i}s M\`{a}rquez, Adam Meyers, Joakim
  Nivre, Sebastian Pad\'{o}, Jan \v{S}t\v{e}p\'{a}nek, Pavel Stra\v{n}\'{a}k,
  Mihai Surdeanu, Nianwen Xue, and Yi~Zhang.
\newblock 2009.
\newblock The {CoNLL}-2009 shared task: Syntactic and semantic dependencies in
  multiple languages.
\newblock In {\em Proceedings of the Thirteenth Conference on Computational
  Natural Language Learning (CoNLL 2009): Shared Task}, pages 1--18, Boulder,
  Colorado, June. Association for Computational Linguistics.

\bibitem[\protect\citename{Hinton and Salakhutdinov}2006]{hinton2006reducing}
Geoffrey~E. Hinton and Ruslan~R. Salakhutdinov.
\newblock 2006.
\newblock Reducing the dimensionality of data with neural networks.
\newblock {\em {S}cience}, 313(5786):504--507.

\bibitem[\protect\citename{Hudson}1984]{hudson1984word}
Richard~A. Hudson.
\newblock 1984.
\newblock {\em Word grammar}.
\newblock Blackwell Oxford.

\bibitem[\protect\citename{Johnson \bgroup et al.\egroup
  }2017]{DBLP:journals/tacl/JohnsonSLKWCTVW17}
Melvin Johnson, Mike Schuster, Quoc~V. Le, Maxim Krikun, Yonghui Wu, Zhifeng
  Chen, Nikhil Thorat, Fernanda~B. Vi{\'{e}}gas, Martin Wattenberg, Greg
  Corrado, Macduff Hughes, and Jeffrey Dean.
\newblock 2017.
\newblock Google's multilingual neural machine translation system: Enabling
  zero-shot translation.
\newblock {\em {TACL}}, 5:339--351.

\bibitem[\protect\citename{Kaiser \bgroup et al.\egroup
  }2017]{DBLP:journals/corr/KaiserGSVPJU17}
Lukasz Kaiser, Aidan~N. Gomez, Noam Shazeer, Ashish Vaswani, Niki Parmar, Llion
  Jones, and Jakob Uszkoreit.
\newblock 2017.
\newblock One model to learn them all.
\newblock {\em CoRR}, abs/1706.05137.

\bibitem[\protect\citename{Kalchbrenner and
  Blunsom}2013]{DBLP:conf/emnlp/KalchbrennerB13}
Nal Kalchbrenner and Phil Blunsom.
\newblock 2013.
\newblock Recurrent continuous translation models.
\newblock In {\em Proceedings of the 2013 Conference on Empirical Methods in
  Natural Language Processing, {EMNLP} 2013, 18-21 October 2013, Grand Hyatt
  Seattle, Seattle, Washington, USA, {A} meeting of SIGDAT, a Special Interest
  Group of the {ACL}}, pages 1700--1709.

\bibitem[\protect\citename{Kingma and Ba}2014]{DBLP:journals/corr/KingmaB14}
Diederik~P. Kingma and Jimmy Ba.
\newblock 2014.
\newblock Adam: {A} method for stochastic optimization.
\newblock {\em CoRR}, abs/1412.6980.

\bibitem[\protect\citename{Kiperwasser and
  Goldberg}2016]{DBLP:journals/tacl/KiperwasserG16}
Eliyahu Kiperwasser and Yoav Goldberg.
\newblock 2016.
\newblock Simple and accurate dependency parsing using bidirectional {LSTM}
  feature representations.
\newblock {\em {TACL}}, 4:313--327.

\bibitem[\protect\citename{Kuncoro \bgroup et al.\egroup
  }2016]{kuncoro-EtAl:2016:EMNLP2016}
Adhiguna Kuncoro, Miguel Ballesteros, Lingpeng Kong, Chris Dyer, and Noah~A.
  Smith.
\newblock 2016.
\newblock Distilling an ensemble of greedy dependency parsers into one mst
  parser.
\newblock In {\em Proceedings of the 2016 Conference on Empirical Methods in
  Natural Language Processing}, pages 1744--1753, Austin, Texas, November.
  Association for Computational Linguistics.

\bibitem[\protect\citename{Kuncoro \bgroup et al.\egroup
  }2017]{kuncoro-EtAl:2017:EACLlong}
Adhiguna Kuncoro, Miguel Ballesteros, Lingpeng Kong, Chris Dyer, Graham Neubig,
  and Noah~A. Smith.
\newblock 2017.
\newblock What do recurrent neural network grammars learn about syntax?
\newblock In {\em Proceedings of the 15th Conference of the European Chapter of
  the Association for Computational Linguistics: Volume 1, Long Papers}, pages
  1249--1258, Valencia, Spain, April. Association for Computational
  Linguistics.

\bibitem[\protect\citename{Lavie and
  Agarwal}2007]{Lavie:2007:MAM:1626355.1626389}
Alon Lavie and Abhaya Agarwal.
\newblock 2007.
\newblock Meteor: An automatic metric for mt evaluation with high levels of
  correlation with human judgments.
\newblock In {\em Proceedings of the Second Workshop on Statistical Machine
  Translation}, StatMT '07, pages 228--231, Stroudsburg, PA, USA. Association
  for Computational Linguistics.

\bibitem[\protect\citename{Linzen \bgroup et al.\egroup }2016]{TACL972}
Tal Linzen, Emmanuel Dupoux, and Yoav Goldberg.
\newblock 2016.
\newblock Assessing the ability of {LSTM}s to learn syntax-sensitive
  dependencies.
\newblock {\em Transactions of the Association for Computational Linguistics},
  4.

\bibitem[\protect\citename{Luong \bgroup et al.\egroup
  }2015a]{DBLP:journals/corr/LuongLSVK15}
Minh{-}Thang Luong, Quoc~V. Le, Ilya Sutskever, Oriol Vinyals, and Lukasz
  Kaiser.
\newblock 2015a.
\newblock Multi-task sequence to sequence learning.
\newblock {\em CoRR}, abs/1511.06114.

\bibitem[\protect\citename{Luong \bgroup et al.\egroup
  }2015b]{DBLP:conf/emnlp/LuongPM15}
Thang Luong, Hieu Pham, and Christopher~D. Manning.
\newblock 2015b.
\newblock Effective approaches to attention-based neural machine translation.
\newblock In {\em Proceedings of the 2015 Conference on Empirical Methods in
  Natural Language Processing, {EMNLP} 2015, Lisbon, Portugal, September 17-21,
  2015}, pages 1412--1421.

\bibitem[\protect\citename{Marcus \bgroup et al.\egroup
  }1993]{DBLP:journals/coling/MarcusSM94}
Mitchell~P. Marcus, Beatrice Santorini, and Mary~Ann Marcinkiewicz.
\newblock 1993.
\newblock Building a large annotated corpus of {E}nglish: The {P}enn treebank.
\newblock {\em Computational Linguistics}, 19(2):313--330.

\bibitem[\protect\citename{Mart\'{i}nez~Alonso and
  Plank}2017]{martinezalonso-plank:2017:EACLlong}
H\'{e}ctor Mart\'{i}nez~Alonso and Barbara Plank.
\newblock 2017.
\newblock When is multitask learning effective? {S}emantic sequence prediction
  under varying data conditions.
\newblock In {\em Proceedings of the 15th Conference of the European Chapter of
  the Association for Computational Linguistics: Volume 1, Long Papers}, pages
  44--53, Valencia, Spain, April. Association for Computational Linguistics.

\bibitem[\protect\citename{Matthews \bgroup et al.\egroup
  }2018]{matthews18naacl}
Austin Matthews, Graham Neubig, and Chris Dyer.
\newblock 2018.
\newblock Using morphological knowledge in open-vocabulary neural language
  models.
\newblock In {\em Meeting of the North American Chapter of the Association for
  Computational Linguistics (NAACL)}, New Orleans, USA, June.

\bibitem[\protect\citename{McDonald \bgroup et al.\egroup
  }2005]{DBLP:conf/acl/McDonaldCP05}
Ryan~T. McDonald, Koby Crammer, and Fernando C.~N. Pereira.
\newblock 2005.
\newblock Online large-margin training of dependency parsers.
\newblock In {\em {ACL} 2005, 43rd Annual Meeting of the Association for
  Computational Linguistics, Proceedings of the Conference, 25-30 June 2005,
  University of Michigan, {USA}}, pages 91--98.

\bibitem[\protect\citename{Melʹ{\v{c}}uk}1988]{melʹvcuk1988dependency}
Igorʹ~Aleksandrovi{\v{c}} Melʹ{\v{c}}uk.
\newblock 1988.
\newblock {\em Dependency {S}yntax: {T}heory and {P}ractice}.
\newblock SUNY press.

\bibitem[\protect\citename{Mikolov \bgroup et al.\egroup }2013]{NIPS2013_5021}
Tomas Mikolov, Ilya Sutskever, Kai Chen, Greg~S. Corrado, and Jeff Dean.
\newblock 2013.
\newblock Distributed representations of words and phrases and their
  compositionality.
\newblock In C.~J.~C. Burges, L.~Bottou, M.~Welling, Z.~Ghahramani, and K.~Q.
  Weinberger, editors, {\em Advances in Neural Information Processing Systems
  26}, pages 3111--3119. Curran Associates, Inc.

\bibitem[\protect\citename{Morishita \bgroup et al.\egroup
  }2017]{DBLP:journals/corr/MorishitaONYSN17}
Makoto Morishita, Yusuke Oda, Graham Neubig, Koichiro Yoshino, Katsuhito Sudoh,
  and Satoshi Nakamura.
\newblock 2017.
\newblock An empirical study of mini-batch creation strategies for neural
  machine translation.
\newblock {\em CoRR}, abs/1706.05765.

\bibitem[\protect\citename{Nadejde \bgroup et al.\egroup
  }2017]{DBLP:journals/corr/NadejdeRSDJKB17}
Maria Nadejde, Siva Reddy, Rico Sennrich, Tomasz Dwojak, Marcin
  Junczys{-}Dowmunt, Philipp Koehn, and Alexandra Birch.
\newblock 2017.
\newblock Syntax-aware neural machine translation using {CCG}.
\newblock {\em CoRR}, abs/1702.01147.

\bibitem[\protect\citename{Neubig \bgroup et al.\egroup }2017]{dynet}
Graham Neubig, Chris Dyer, Yoav Goldberg, Austin Matthews, Waleed Ammar,
  Antonios Anastasopoulos, Miguel Ballesteros, David Chiang, Daniel Clothiaux,
  Trevor Cohn, Kevin Duh, Manaal Faruqui, Cynthia Gan, Dan Garrette, Yangfeng
  Ji, Lingpeng Kong, Adhiguna Kuncoro, Gaurav Kumar, Chaitanya Malaviya, Paul
  Michel, Yusuke Oda, Matthew Richardson, Naomi Saphra, Swabha Swayamdipta, and
  Pengcheng Yin.
\newblock 2017.
\newblock Dy{N}et: {T}he dynamic neural network toolkit.
\newblock {\em arXiv preprint arXiv:1701.03980}.

\bibitem[\protect\citename{Niehues and Cho}2017]{DBLP:conf/wmt/NiehuesC17}
Jan Niehues and Eunah Cho.
\newblock 2017.
\newblock Exploiting linguistic resources for neural machine translation using
  multi-task learning.
\newblock In {\em Proceedings of the Second Conference on Machine Translation,
  {WMT} 2017, Copenhagen, Denmark, September 7-8, 2017}, pages 80--89.

\bibitem[\protect\citename{Papineni \bgroup et al.\egroup
  }2002]{DBLP:conf/acl/PapineniRWZ02}
Kishore Papineni, Salim Roukos, Todd Ward, and Wei{-}Jing Zhu.
\newblock 2002.
\newblock {BLEU}: a method for automatic evaluation of machine translation.
\newblock In {\em Proceedings of the 40th Annual Meeting of the Association for
  Computational Linguistics, July 6-12, 2002, Philadelphia, PA, {USA.}}, pages
  311--318.

\bibitem[\protect\citename{Reimers and
  Gurevych}2017]{DBLP:conf/emnlp/ReimersG17}
Nils Reimers and Iryna Gurevych.
\newblock 2017.
\newblock Reporting score distributions makes a difference: Performance study
  of {LSTM}-networks for sequence tagging.
\newblock In {\em Proceedings of the 2017 Conference on Empirical Methods in
  Natural Language Processing, {EMNLP} 2017, Copenhagen, Denmark, September
  9-11, 2017}, pages 338--348.

\bibitem[\protect\citename{Rush \bgroup et al.\egroup
  }2015]{DBLP:conf/emnlp/RushCW15}
Alexander~M. Rush, Sumit Chopra, and Jason Weston.
\newblock 2015.
\newblock A neural attention model for abstractive sentence summarization.
\newblock In {\em Proceedings of the 2015 Conference on Empirical Methods in
  Natural Language Processing, {EMNLP} 2015, Lisbon, Portugal, September 17-21,
  2015}, pages 379--389.

\bibitem[\protect\citename{Sennrich and Haddow}2016]{DBLP:conf/wmt/SennrichH16}
Rico Sennrich and Barry Haddow.
\newblock 2016.
\newblock Linguistic input features improve neural machine translation.
\newblock In {\em Proceedings of the First Conference on Machine Translation,
  {WMT} 2016, colocated with {ACL} 2016, August 11-12, Berlin, Germany}, pages
  83--91.

\bibitem[\protect\citename{Sennrich \bgroup et al.\egroup
  }2015]{DBLP:journals/corr/SennrichHB15}
Rico Sennrich, Barry Haddow, and Alexandra Birch.
\newblock 2015.
\newblock Neural machine translation of rare words with subword units.
\newblock {\em CoRR}, abs/1508.07909.

\bibitem[\protect\citename{Sennrich \bgroup et al.\egroup
  }2016]{DBLP:conf/acl/SennrichHB16a}
Rico Sennrich, Barry Haddow, and Alexandra Birch.
\newblock 2016.
\newblock Neural machine translation of rare words with subword units.
\newblock In {\em Proceedings of the 54th Annual Meeting of the Association for
  Computational Linguistics, {ACL} 2016, August 7-12, 2016, Berlin, Germany,
  Volume 1: Long Papers}.

\bibitem[\protect\citename{S{\o}gaard and Goldberg}2016]{sogaard2016deep}
Anders S{\o}gaard and Yoav Goldberg.
\newblock 2016.
\newblock Deep multi-task learning with low level tasks supervised at lower
  layers.
\newblock In {\em Proceedings of the 54th Annual Meeting of the Association for
  Computational Linguistics (Volume 2: Short Papers)}, volume~2, pages
  231--235.

\bibitem[\protect\citename{Sutskever \bgroup et al.\egroup
  }2014]{DBLP:conf/nips/SutskeverVL14}
Ilya Sutskever, Oriol Vinyals, and Quoc~V. Le.
\newblock 2014.
\newblock Sequence to sequence learning with neural networks.
\newblock In {\em Advances in Neural Information Processing Systems 27: Annual
  Conference on Neural Information Processing Systems 2014, December 8-13 2014,
  Montreal, Quebec, Canada}, pages 3104--3112.

\bibitem[\protect\citename{Vinyals \bgroup et al.\egroup
  }2015]{DBLP:conf/nips/VinyalsKKPSH15}
Oriol Vinyals, Lukasz Kaiser, Terry Koo, Slav Petrov, Ilya Sutskever, and
  Geoffrey~E. Hinton.
\newblock 2015.
\newblock Grammar as a foreign language.
\newblock In {\em Advances in Neural Information Processing Systems 28: Annual
  Conference on Neural Information Processing Systems 2015, December 7-12,
  2015, Montreal, Quebec, Canada}, pages 2773--2781.

\bibitem[\protect\citename{Wang \bgroup et al.\egroup
  }2015]{DBLP:journals/corr/WangQSHZ15}
Peilu Wang, Yao Qian, Frank~K. Soong, Lei He, and Hai Zhao.
\newblock 2015.
\newblock Part-of-speech tagging with bidirectional long short-term memory
  recurrent neural network.
\newblock {\em CoRR}, abs/1510.06168.

\bibitem[\protect\citename{Yosinski \bgroup et al.\egroup
  }2014]{DBLP:conf/nips/YosinskiCBL14}
Jason Yosinski, Jeff Clune, Yoshua Bengio, and Hod Lipson.
\newblock 2014.
\newblock How transferable are features in deep neural networks?
\newblock In {\em Advances in Neural Information Processing Systems 27: Annual
  Conference on Neural Information Processing Systems 2014, December 8-13 2014,
  Montreal, Quebec, Canada}, pages 3320--3328.

\bibitem[\protect\citename{Zhang \bgroup et al.\egroup
  }2017]{zhang-EtAl:2017:EMNLP20173}
Zhirui Zhang, Shujie Liu, Mu~Li, Ming Zhou, and Enhong Chen.
\newblock 2017.
\newblock Stack-based multi-layer attention for transition-based dependency
  parsing.
\newblock In {\em Proceedings of the 2017 Conference on Empirical Methods in
  Natural Language Processing}, pages 1678--1683, Copenhagen, Denmark,
  September. Association for Computational Linguistics.

\bibitem[\protect\citename{Zoph and Knight}2016]{DBLP:conf/naacl/ZophK16}
Barret Zoph and Kevin Knight.
\newblock 2016.
\newblock Multi-source neural translation.
\newblock In {\em {NAACL} {HLT} 2016, The 2016 Conference of the North American
  Chapter of the Association for Computational Linguistics: Human Language
  Technologies, San Diego California, USA, June 12-17, 2016}, pages 30--34.

\end{thebibliography}
\bibliographystyle{acl2012}

\end{document}